%% file: emnlp2022.tex
\pdfoutput=1
\documentclass[11pt]{article}

\usepackage{EMNLP2022}

\usepackage{times}
\usepackage{latexsym}
\usepackage{booktabs}
\usepackage{comment}

\usepackage[T1]{fontenc}
\usepackage[utf8]{inputenc}

\usepackage{microtype}

\usepackage[pdftex]{graphicx}
\title{T-Modules: Translation Modules\\ for Zero-Shot Cross-Modal Machine Translation}

\author{
  \hspace*{40pt} Paul-Ambroise Duquenne \\
  \hspace*{40pt} Meta AI \& Inria \\
  \hspace*{40pt} \texttt{\small padqn@fb.com} \\
  \And
  \hspace*{37pt} Hongyu Gong \\
  \hspace*{37pt} Meta AI \\
  \hspace*{37pt} \texttt{\small hygong@fb.com} \\
  \And
  \hspace*{-25pt} Benoît Sagot \hspace*{-40pt}\\
  \hspace*{-25pt} Inria \hspace*{-40pt}\\
  \hspace*{-25pt} \texttt{\small benoit.sagot@inria.fr} \hspace*{-40pt} \\
  \And
  \hspace*{-15pt} Holger Schwenk \\
  \hspace*{-15pt} Meta AI \\
  \hspace*{-15pt} \texttt{\small schwenk@fb.com}
 }

\begin{document}

\newcommand{\ttot}{\texttt{t2t}}
\newcommand{\stot}{\texttt{s2t}}
\newcommand{\stos}{\texttt{s2s}}
\newcommand{\ttos}{\texttt{t2s}}

\newcommand{\MC}{\multicolumn}
\newcommand{\MR}{\multirow}
\newcommand{\LB}{\linebreak[4]}

\newcommand{\holger}[1]{{\color{orange}\textbf{Holger:} #1}}
\newcommand{\paul}[1]{{\color{blue!20!red}[\textbf{PA:} #1]}}
\newcommand{\hongyu}[1]{{\color{blue}[\textbf{Hongyu:} #1]}}
\newcommand{\benoit}[1]{{\color{green}[\textbf{Benoît:} #1]}}
\newcommand{\todo}[1]{{\color{red}[\textbf{TODO:} #1]}}

\maketitle
\begin{abstract}
\input{0-abstract}

\end{abstract}

\input{1-introduction}
\input{2-related_work}

\input{3-motivation}

\input{4-model}
\input{5-conclusion}

\bibliography{anthology,custom}
\bibliographystyle{acl_natbib}
\newpage
\appendix
\input{6-appendix}
\end{document}

%% file: 0-abstract.tex
We present a new approach to perform zero-shot cross-modal transfer between speech and text for translation tasks. Multilingual speech and text are encoded in a joint fixed-size representation space. Then, we compare different approaches to decode these multimodal and multilingual fixed-size representations, enabling zero-shot translation between languages and modalities.
All our models are trained without the need of cross-modal labeled translation data.
Despite a fixed-size representation, we achieve very competitive results on several text and speech translation tasks. In particular, we outperform the state of the art for zero-shot speech translation on Must-C. \iffalse Incorporating a speech decoder in our framework, we\fi{} We also introduce the first results for zero-shot direct speech-to-speech and text-to-speech translation.

%% file: 1-introduction.tex
\section{Introduction}

Most, if not all, current state-of-the-art text and speech translation systems are based on a sequence-to-sequence approach and an attention mechanism to connect the encoder and decoder.
Such models require labeled data to be trained end-to-end.
For text-to-text (T2T) translation, this labeled data, called bitexts, is available in large amounts for a number of language pairs, in particular since large-scale bitext mining initiatives like ParaCrawl \cite{banon-etal-2020-paracrawl} and CCMatrix \cite{schwenk:2021:acl_ccmatrix}. Finding training data for speech-to-text (S2T) translation is more challenging, but several data collection efforts exist, like mTEDx \cite{salesky2021multilingual}, CoVoST \cite{covost1:2020:lrec,covost2:2020:arxiv}, and Must-C \cite{mustc:2019:naacl}.
Finally, speech-to-speech (S2S) translation suffers from scarcity of end-to-end labeled data and current S2S systems are limited to a very small number of language pairs. Very recent works start to consider mining labeled data for S2S, e.g. \cite{Duquenne:2021:nips_mine}.

Unsupervised representation learning is very successfully used to initialize the encoder and/or decoder of a sequence-to-sequence model, thereby lowering the amount of labeled data needed to train or fine-tune the model end-to-end. Approaches include for instance XLM \cite{xlm},
XLSR \cite{xlsr},
wav2vec \cite{wav2vec2},
data2vec \cite{baevski:2022:arxiv_data2vec}
and mSLAM \cite{mslam}.

\begin{figure}
\begin{center}
    \includegraphics[width=1\linewidth]{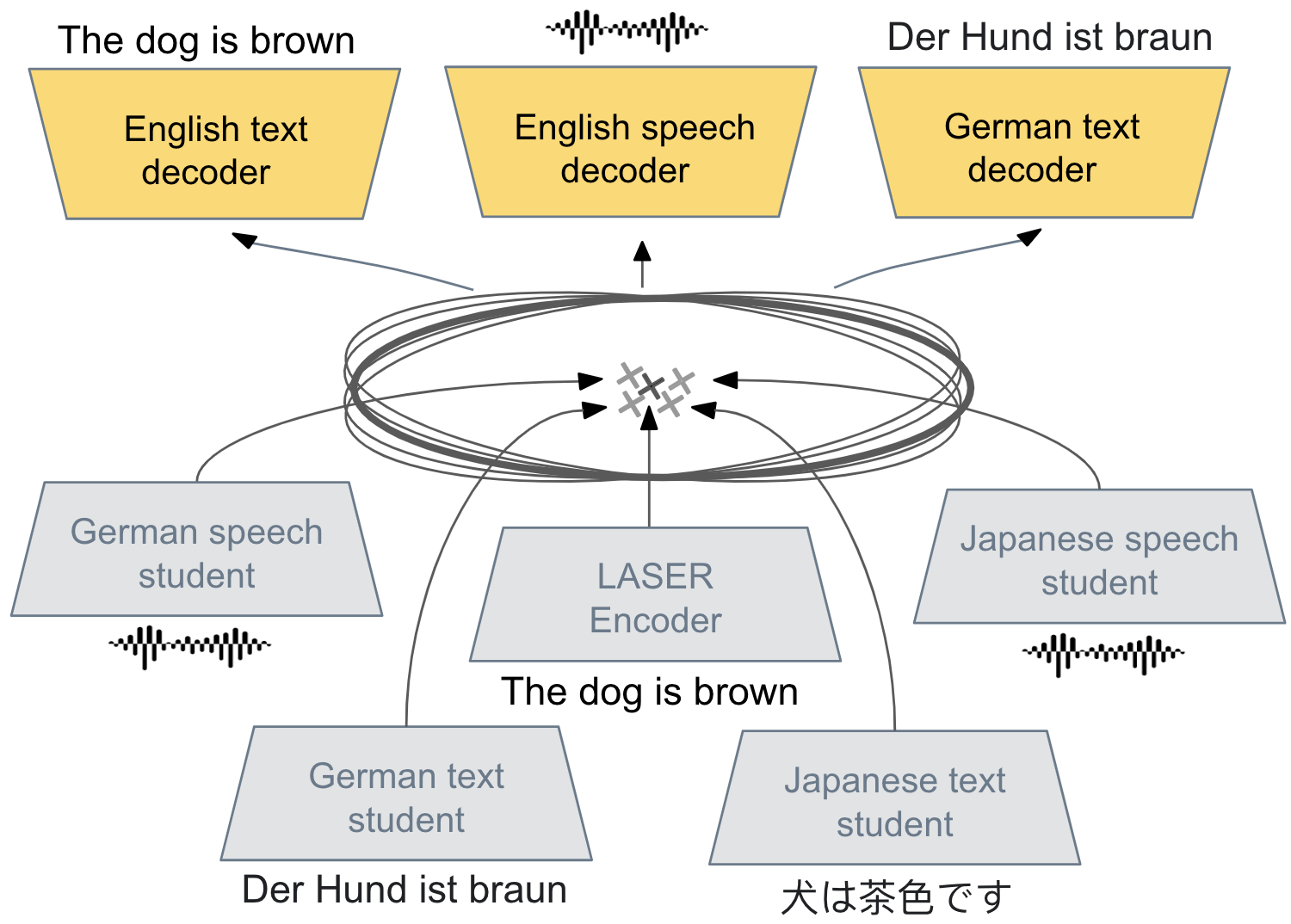}
  \caption{Summary of the model architecture.}
  \label{fig:arch}
\end{center}
\end{figure}

In this work, we propose a new modular architecture for text and speech translation, which is based on a \textbf{common fixed-size multilingual and multimodal internal representation}, and encoders and decoders which are independently trained. 
We explore several variants of teacher-student training to learn text and speech encoders for multiple languages, which are compatible with the embedding space of the LASER encoder \cite{Artetxe:2019:tacl_massive_ml}. \iffalse\footnote{\url{https://github.com/facebookresearch/LASER}}\fi{} 
In contrast to preceding works on multilingual and multimodal representations, we also train text decoders for multiple languages which are able to generate translations given the joint representation. Finally, we demonstrate that it is possible to train a speech decoder using raw audio only. Figure~\ref{fig:arch} visualizes the overall approach.
We show that these encoders and decoders can be freely combined to achieve very competitive performance in T2T, S2T and (zero-shot) S2S translation.

\begin{comment}
In summary, our main contributions are:
(1)~We apply a teacher-student approach to train multilingual text and speech encoders that are mutually compatible;
(2)~We show that the fixed-size representation can be efficiently decoded into multiple languages;
(3)~We are able to train a speech decoder with raw speech only, which can be paired with our text and speech encoders for multiple languages;
(4)~We achieve very competitive results on several text and speech translation tasks, without any end-to-end labeled data and significantly improve the state of the art for zero-shot speech translation;
 and (5)~To the best of our knowledge, we are the first to build zero-shot direct S2S translation systems.
\end{comment}

In summary, our contributions are as follows.
\begin{itemize}
    \item We apply a teacher-student approach to train multilingual text and speech encoders that are mutually compatible;
    \item We show that the fixed-size representation can be efficiently decoded into multiple languages;
    \item We are able to train a speech decoder with raw speech only, which can be paired with our text and speech encoders for multiple languages;
    \item We achieve very competitive results on several text and speech translation tasks, without any end-to-end labeled data and significantly improve the state of the art for zero-shot speech translation;
    \item To the best of our knowledge, we are the first to build zero-shot direct S2S translation systems.
\end{itemize}

\iffalse
This paper is structured as follows. In the next section, we first summarize the related work. We then explore several training strategies for the encoders and decoders. Based on this initial study, we extend our techniques to multiple languages in the text and speech modalities (Section~\ref{sec:archi}). Main results are summarized in Section~\ref{sec:results}, and the paper concludes with a discussion. 
\fi{}

%% file: 2-related_work.tex
\section{Related work}
\paragraph{Multilingual and multimodal representations}
Building multilingual representation for text or speech is key to develop state-of-the-art models based on these modalities.  \citet{xlm} introduce a multilingual pre-training method with good cross-lingual transfer capabilities. \citet{xlsr} extend the Wav2vec2 \citep{wav2vec2} architecture to the multilingual setting introducing a multilingual pre-trained model for speech. More recently, \citet{mslam} pre-train a multilingual encoder model handling both speech and text in order to benefit from cross-modal transfer between speech and text.

An important obstacle to good joint speech/text representations is the length mismatch between audio and text. On the other hand, several works have studied how to encode sentences in a fixed-size representation \cite{labse, Artetxe:2019:tacl_massive_ml,sentence_bert}. In the multilingual setting, these works highlight that paraphrases and translations are close in the sentence embedding space, enabling large-scale bitext mining. Recently, \citet{Duquenne:2021:nips_mine} extended the existing LASER model \citep{Artetxe:2019:tacl_massive_ml} built for multilingual text to the speech modality for several spoken languages. They show that this joint speech/text fixed-size representation can be efficiently used for large-scale mining of speech against text and even speech against speech.

\paragraph{Zero-shot transfer in Machine Translation}
In Machine Translation, cross-lingual transfer to improve low-resource language directions has been widely studied. One way to encourage cross-lingual transfer is building a massively multilingual translation system as \citep{m2m100}. Some other works such as \citep{zhang2022triangular} make an efficient use of MT data involving a pivot language thanks to weight freezing strategies to force representations to be close to the pivot language representations. One extreme scenario of cross-lingual transfer learning is called zero-shot transfer, where you learn to translate one language and directly apply the decoding process to an unseen language. Several methods have been tried to improve zero-shot transfer. \citet{arivazhagan2019missing,pham2019improving} add language similarity regularization on pooled representations of encoders outputs as an auxiliary loss to a MT objective in order to improve zero-shot transfer. \citet{liao2021improving,vazquez2018multilingual,lu2018neural} introduce shared weights between language-specific encoders and decoders, commonly called an interlingua that captures language-independent semantic information. Finally, \citet{escolano2020multilingual,escolano2021bilingual,escolano2020training} focus on incremental learning of language-specific encoders-decoders using cross-entropy loss, alternately freezing parts of the model to ensure a shared representation between languages.

\paragraph{Zero-shot transfer in Speech Translation}
Recent research focuses on direct speech translation where an encoder-decoder model directly translates speech into text \citep{berard2016listen,bansal2017towards,weiss2017sequence}. Direct speech translation models are closing the gap with their cascaded counterparts \citep{li2020multilingual,babu2021xls,mslam}. Several works add MT data in S2T translation training, using an auxiliary loss to bridge the modality gap, like adversarial \cite{alinejad2020effectively}, or distance \cite{dong2021listen,liu2020bridging} regularization. \cite{xu2021stacked} and \cite{li2020multilingual} use adaptor modules to address the length mismatch between audio and text representations. Several works studied how to efficiently perform zero-shot cross-modal transfer from text to speech in the frame of direct speech translation.
Following \citep{escolano2020multilingual,escolano2021bilingual,escolano2020training} presented above for text, \citeauthor{escolano2021enabling} learn a speech encoder compatible with decoders trained on text only, freezing the text decoder during training and using cross-entropy on the output of the decoder. This is the most similar work like ours, however they did not use any joint fixed-representation and their zero-shot results using only speech transcriptions lagged behind supervised setting by a large margin.
Other works such as \citep{dinh2022tackling,dinh2021zero} studied zero-shot speech translation employing a cross-modal similarity regularization as an auxiliary loss. However, they obtained low zero-shot results possibly due to the mismatch in the encoder output lengths between speech and text.

\paragraph{Direct speech-to-speech translation}
Finally, there is a surge of research interest in direct speech-to-speech translation \citep{jia2019direct,jia2021translatotron,s2s_paper}. An encoder-decoder model directly translates speech in a language into speech in another language without the need to generate text as an intermediate step. Speech-to-speech translation research suffers from data scarcity of aligned speech with speech in different languages and often uses synthetic speech to overcome this issue. 

Recently, \citet{s2s_real_data} introduce the first direct speech-to-speech model based on real speech data as target. They propose a speech normalization technique in order to normalize the target speech with respect to speaker and prosody. \citet{s2s_paper,s2s_real_data} extract HuBERT units of target speech as targets for a unit decoder during training. At test time, a vocoder is used to transform output units into speech.
To the best of our knowledge, no work has tried to develop a direct speech-to-speech translation system in a zero-shot setting.

%% file: 3-motivation.tex
\section{Exploring training strategies}

\begin{comment}
\begin{figure}[t!]
    \centering
    \includegraphics[width=0.75\linewidth]{images/autoencoding.png}
    \caption{Auto-encoding evaluation of English sentences from FLORES.}
    \label{fig:autoencoding}
\end{figure}

\begin{figure}[t!]
    \centering
    \includegraphics[width=0.75\linewidth]{images/translation.png}
    \caption{Translation evaluation for en-de direction from FLORES.}
    \label{fig:translation}
\end{figure}
\end{comment}

The purpose of this work is to build a common fixed-size representation for multilingual speech and multilingual text that can be decoded in text and speech in different languages. We want to build language-specific encoders and decoders compatible with this fixed-size representation. Plugging one encoder with one decoder from different modalities and/or different languages enables performing zero-shot cross-modal translation. 

To this end, we first study how to efficiently decode fixed-size sentence representation for text. Second, we study how to improve similarity for sentence embeddings between languages. After an ablation study on the Japanese-English text translation direction, we extend the best training strategy to several other languages and a new modality, speech.

\begin{figure}[t]
    \centering
    \includegraphics[width=1\linewidth]{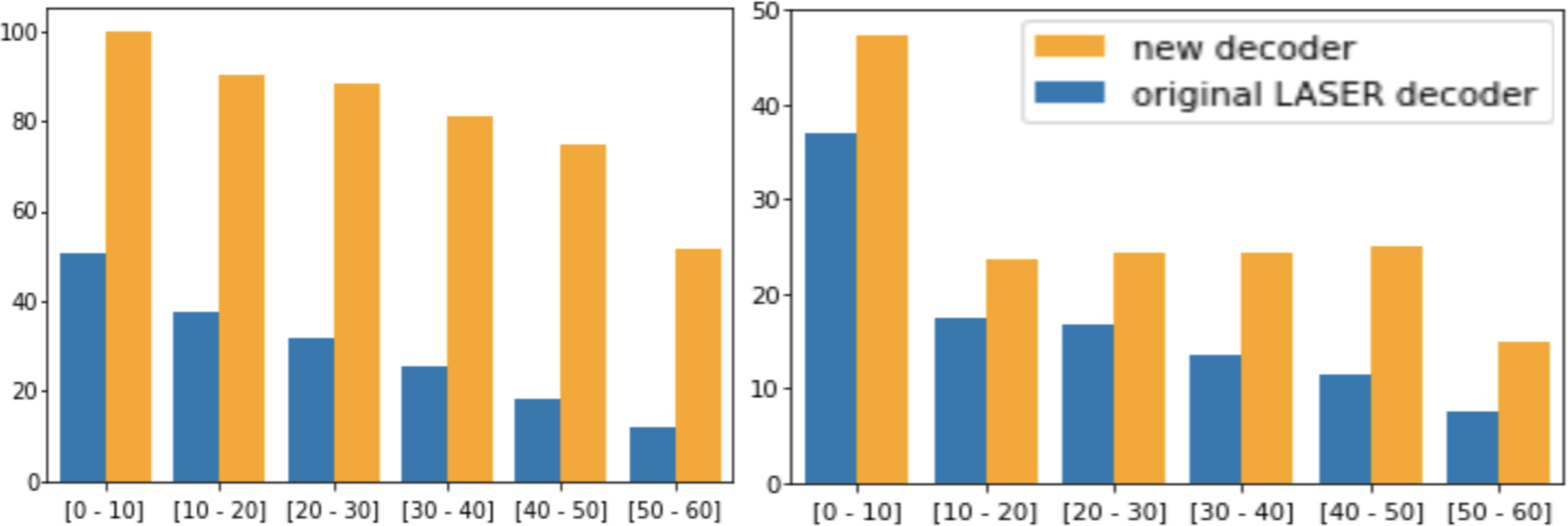}
    \caption{BLEU vs.~sentence length on FLORES-devtest. English auto-encoding (left), German-to-English translation (right).}
    \label{fig:auto_trans}
\end{figure}

\subsection{Better decoding of sentence embeddings}
\paragraph{Motivations} Multilingual sentence embeddings have been widely studied in the research community to perform bitext mining. For instance, LASER \citep{Artetxe:2019:tacl_massive_ml} is a multilingual sentence embedding space, where sentences are close in the embedding space if they are paraphrases or translations. LASER has been successfully used for large-scale bitext mining like in the CCMatrix project \citep{schwenk:2021:acl_ccmatrix}. LASER has been trained with a decoding objective, whereas other works like LaBSE \citep{labse} have been trained with a contrastive objective.

First, we studied how multilingual sentence embeddings can be efficiently decoded. We focused on LASER as it originally has a decoder, and we studied how we can improve the decoding of sentence embeddings. As an initial experiment, we evaluated auto-encoding of English sentences from FLORES \citep{flores} in \autoref{fig:auto_trans} left, %\ref{fig:autoencoding},
with the original LASER encoder and decoder, bucketing sentences by length, and reporting BLEU scores. 
The LASER encoder handles several languages: decoding these multilingual embeddings enables to translate the input sentence into English with the original LASER decoder. We report the BLEU scores for the different sentence lengths in \autoref{fig:auto_trans} right %\ref{fig:translation}
for the German-English translation direction from FLORES. We notice that BLEU scores are low for both auto-encoding and translation tasks and decrease with the sentence length. The fixed-size representation seems to be a bottleneck for decoding tasks, especially for long sentences. However, the original LASER decoder is really shallow (one LSTM decoder layer), an interesting question is: can we improve decoding by training a new deeper decoder?

\paragraph{Training new decoders} We chose to train a new decoder to decode LASER sentence embeddings, with a transformer architecture and 12 layers. To train this new decoder, we use an auto-encoding objective, feeding raw English sentences to the model: we use original LASER encoder, whose weights are not updated during training, and plug a new transformer decoder to decode the fixed-size sentence representation output by the LASER encoder (the decoder attends on the sentence embedding output by the encoder). We used 15B English sentences from CCnet \citep{wenzek2019ccnet} to train the decoder. We compare the new decoder with original LASER decoder on the auto-encoding task and the German-English translation task of FLORES in \autoref{fig:auto_trans}. %\ref{fig:autoen_coding} and Figure~\ref{fig:translation}. 

\paragraph{Results} First, we notice an important boost on the auto-encoding task with the new decoder, with high BLEU scores even for sentences with more than 50 words. Second, training a new decoder with an auto-encoding objective improves the decoding of sentence embeddings from another language, German. The new decoder can be directly applied to German sentence embeddings because German embeddings are supposed to be close to their English translations encoded with LASER.

\subsection{Making languages closer}
\paragraph{Motivations} To get an idea of the closeness of translations in the LASER space, we inspected the L2 squared distances of sentence embeddings in different languages to their English translations sentence embeddings. A detailed analysis can be found in the appendix. We noticed that high resource languages are closer in the LASER space to English, compared to low resource languages. 
%Figure~\ref{fig:distances} highlights that Japanese translations are more distant to English translations compared to German translations. 

We studied how our newly trained decoder is performing on a more distant language in LASER space, Japanese. We report the results of the ja-en translation task using the original decoder and the new decoder in \autoref{tab:ja_en}. We notice that both decoders performs poorly on the ja-en translation tasks, and that the original LASER decoder leads to better results. An hypothesis is that the new decoder has over-fitted English embeddings leading to bad generalization on distant Japanese embeddings.

\paragraph{Teacher-student training of text encoders} To overcome this issue, we suggest to follow a method introduced by \citet{Reimers}, where new encoders are trained to fit an existing sentence embedding space. Here, we are trying to make the Japanese translations closer to English embeddings in our 1024 dimensional space. The original LASER encoder is fixed during training to encode English translation, behaving as the teacher, while we train a new Japanese encoder as a student to fit English sentence embeddings. We use bitexts from CCMatrix for the ja-en pair to train the Japanese text student. Following \citep{Reimers}, we minimize the MSE loss (equivalent to L2 squared distance) between the generated Japanese sentence embedding and the target English sentence embedding. 

The Japanese encoder is not trained from scratch, but we fine-tune XLM-R large. To extract the sentence embedding, we tested two methods: The classical output of the encoder corresponding to the beginning-of-sentence (BOS) token, a method widely used for text classification ; or max-pooling of the encoder outputs, less common but LASER has been trained with such pooling method. 

Finally, we tested another objective that is supposed to better match with our decoding task: we encode the Japanese sentence with the encoder being trained, decode the pooled sentence embedding with our new decoder which weights are not updated during training, and we compute the cross entropy loss of the output of the new decoder with the English target sentence. The training was unstable when using XLM-R weights as initialization. Therefore, instead of fine-tuning XLM-R, we fine-tune the encoder obtained from our previous method (trained with MSE loss), which leads to a stable training. We report all the results in \autoref{tab:ja_en}. For text-to-text translation results, we use spBLEU of M2M-100 with the public checkpoint and script to evaluate on FLORES.

\begin{table}[!t]
    \centering\small
    \begin{tabular}{lr}
    \toprule
        ~ & ja-en \\ \midrule
        Original encoder + original decoder & 6.9 \\ 
        Original encoder + new decoder & 5.5 \\ 
        Student - BOS pooling + new decoder & 19.5 \\ 
        Student - max pooling  + new decoder & 22.5 \\ 
        Student - max pooling  + original decoder & 12.2 \\ 
        Student - max pooling \& CE + new decoder & 22.6 \\ \bottomrule
    \end{tabular}
    \caption{BLEU scores for ja-en on FLORES devtest}
        \label{tab:ja_en}
\end{table}

\paragraph{Results} In \autoref{tab:ja_en}, we first notice that learning a new Japanese student significantly improve the results for the ja-en translation task. The best pooling method seems to be max-pooling, maybe because LASER has been trained with max-pooling. The second step of fine-tuning with cross entropy loss does not improve the results for our ja-en translation task, despite of the significant decrease of cross entropy valid loss during this second step fine-tuning. This validates the use a simple MSE loss which seems sufficient for future decoding purposes and is a lot cheaper in term of computation compared to cross entropy loss. We conclude that learning a new Japanese student with max-pooling and MSE loss leads to the best results. Using this new Japanese encoder, our new decoder significantly outperforms the original LASER encoder.

These experiments show that LASER sentence embeddings can be better decoded by training a new decoder on a large amount of raw text data. This new decoder can be used to decode sentence embeddings from other languages handled by LASER. However, translations are still more or less distant in the space, making them explicitly closer with a MSE loss objective significantly improves the results on a translation task. Therefore, we decide to extend this idea to other languages and a new modality, speech, to see if it can help performing cross-modal translation tasks.

%% file: 4-model.tex
\section{Overall architecture}
\label{sec:archi}

\begin{table*}[!t]
    \centering\small
    \begin{tabular}{l*{7}{r}}
    \toprule
         & \MC{1}{c}{de} & \MC{1}{c}{fr} & \MC{1}{c}{es} & \MC{1}{c}{ca} & \MC{1}{c}{ja} & \MC{1}{c}{tr} & \MC{1}{c}{mn} \\ \midrule
%         \multicolumn{8}{l}{\textbf{Translation into English}} \\
        \MC{8}{l}{\textbf{\textit{~~~This work - zero-shot except for de-en}}} \\
        ~~~en-en decoder & 40.7 & 41.9 & 30.4 & 36.7 & 22.5 & 32.8 & 13.0 \\ 
        ~~~en-en+noise decoder & 39.5 & 40.6 & 29.4 & 35.8 & 23.7 & 33.2 & 16.4 \\ 
       ~~~en-en+de-en decoder & 44.2 & 44.9 & 32.6 & 40.7 & 26.5 & 37.3 & 19.4 \\ 
        \MC{8}{l}{\textbf{\textit{~~~Previous works - supervised}}} \\
        ~~~M2M-100 (12B - 48 layers) \citep{m2m100} & 44.7 & 45.5 & 31.1 & 42.5 & 26.1 & 36.9 & 20.9 \\
        ~~~Deepnet (3.2B - 200 layers) \citep{wang2022deepnet} & 48.0 & 49.9 & 35.2 & 46.2 & 32.7 & 44.2 & 23.9 \\\bottomrule
    \end{tabular}
     
    \caption{BLEU on FLORES devtest for text-to-text xx-en translation using different English decoders.}
    \label{tab:t2t_en}
\end{table*}

%\subsection{Text modules}
%In this part, we present the overall architecture of our model.
\paragraph{Text student encoders}
We now want to train several text students for different languages, in order to plug, at test time, these  encoders to different decoders to perform translation tasks. 
We decide to use LASER \textbf{English} embeddings as our teacher. This English space has proven to have good semantic properties: paraphrases are close in the embedding space, and makes it a good teacher for English translations. Moreover, most of MT data involve English translations that we will use to learn our text students.
We focus on 7 languages, namely, German, French, Spanish, Catalan, Japanese, Turkish, and Mongolian. We use CCMatrix bi-texts to learn our text students, and bi-texts mined with LASER3 \cite{laser3} for Mongolian.
\paragraph{Text decoders}
We saw above that we can train a new English decoder with raw English data, using a fixed encoder and an auto-encoding objective. However, such an approach can lead to over-fitting to English sentence embeddings and bad generalization on other languages. We made languages closer together in our 1024 dimensional space thanks to our new student encoders but translations are not perfectly mapped to a real English sentence embedding in this continuous space. Therefore, we explore different methods to make the decoders robust locally in the sentence embedding space in order to generalize better on unseen languages.

First, we can improve our decoder training with an auto-encoding objective by adding synthetic noise in the sentence embedding space. We add noise to a sentence embedding by multiplying it by $1 + \epsilon$, with $\epsilon \sim \mathcal{N}(0,\alpha)$.
In our experiments, we took $\alpha=0.25$, which leads to an empirical average L2 squared distance of approx.~$0.05$. between the noisy embedding and the original embedding.

Second, we tested another approach to make our decoder robust to translations in the sentence embedding space: we added bi-texts from the de-en direction to the training of the English decoder. 

Finally, we trained decoders for five non-English languages to see how it behaves for other languages. All text decoders are 12-layers transformer decoders.

\begin{figure}[t!]
    \centering
    \includegraphics[width=1\linewidth]{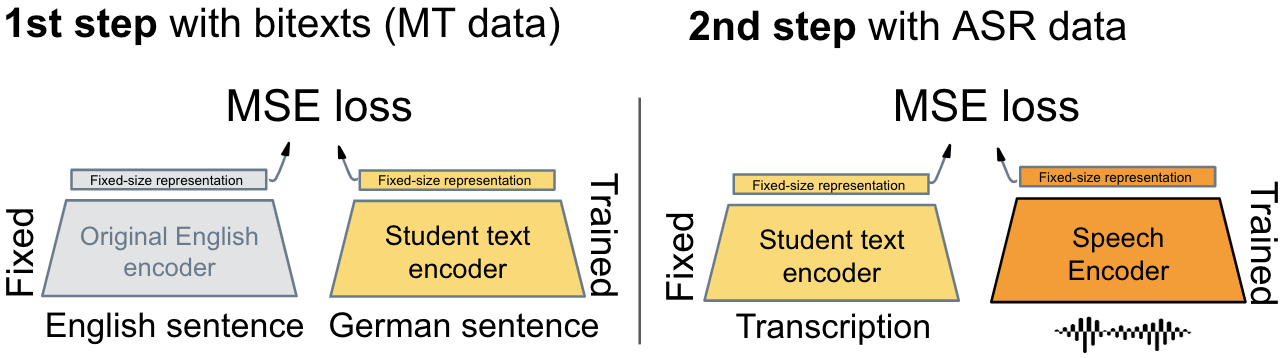}
    \caption{Incremental learning of a speech student.}
    \label{fig:speech_student}
\end{figure}

%\subsection{Speech modules}
\paragraph{Speech student encoders}
\citet{Duquenne:2021:nips_mine} showed that it is possible to learn speech students compatible with the LASER text space. The training of speech students is similar to the one presented above for text. They fine-tune XLSR, a multilingual pretrained model for speech and minimized the cosine loss between the output of the speech encoder and the target LASER sentence embedding.
We adapt this approach using a bigger XLSR model \citep{babu2021xls} with more than two billion parameters and extracting the fixed-size representation for speech with max-pooling to follow what we have done for text students. We minimize the MSE loss between the output of the speech encoder and the transcription/translation encoded by one of our text encoders. Unlike \citep{Duquenne:2021:nips_mine}, we did not use the original LASER encoder to encode text transcripts but our newly trained text students which are supposed to be close to the LASER English embeddings.
As in \citep{Duquenne:2021:nips_mine}, we can use either transcriptions or written translations as teachers for our speech student. We used CoVoST 2, a speech translation dataset, as our training data.
\autoref{fig:speech_student} summarizes the process to train a speech student with transcriptions only: First, we train a text student for the language we want to cover, we will use this encoder to encode transcriptions. Then, we train a speech student to fit text embeddings output by our text student.

\begin{figure}[b!]
    \centering
    \includegraphics[width=0.6\linewidth]{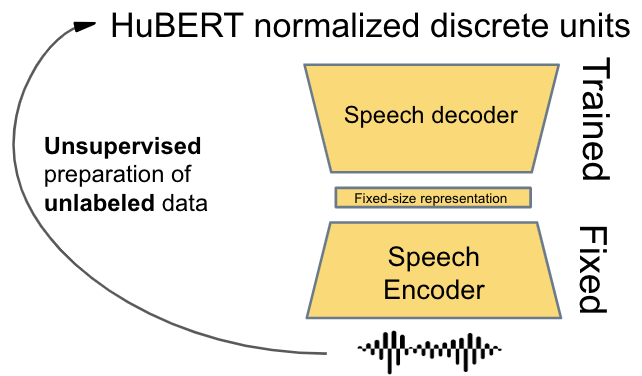}
    \caption{Speech decoder training.}
    \label{fig:speech_decoder}
\end{figure}

\paragraph{Speech decoders}
In this last part, we introduce a speech decoder in our framework, which can be learnt with raw speech data. We focus on English speech decoding but it could be extended to other languages. To learn to decode English speech, we follow the work done by \citet{s2s_real_data}, who learn to decode HuBERT units. At test time, the generated units are transformed into speech using a vocoder.

One method is to follow the same approach presented for raw text data to learn an English decoder. The English speech encoder previously trained to fit LASER text space on CoVoST 2 training set is used to encode raw speech, and its weights are not updated during training. We trained a unit decoder to decode sentence embeddings output by the speech encoder. The unit targets correspond to the one of the input speech as we are trying to auto-encode speech. We follow the recipe of \citet{s2s_real_data} to prepare target units as we are dealing with real speech data: we extract HuBERT units from input speech, normalize the units with the speech normalizer used in \citet{s2s_real_data}. This preparation of target data is done unsupervisedly and any raw speech data can be processed with this method. We summarize the speech decoder training in \autoref{fig:speech_decoder}.
Another method is to leverage English speech recognition data where English text transcripts are encoded through LASER encoder which weights are fixed during training and a decoder predicts the sequence of units of the corresponding speech.

Once the English speech decoder is trained, we can plug any text or speech encoder to perform direct text-to-speech or speech-to-speech translation in a zero-shot way.

\section{Results and discussion}
\label{sec:results}

\begin{table*}[!ht]
    \centering\small
    \begin{tabular}{l*{7}{r}}
    \toprule
         & \MC{1}{c}{de} & \MC{1}{c}{fr} & \MC{1}{c}{es} & \MC{1}{c}{ca} & \MC{1}{c}{tr} & \MC{1}{c}{ja} & \MC{1}{c}{mn} \\ \midrule
        Speech training hours in CoVoST 2 & 184h & 264h & 113h & 136h & 4h & 2h & 3h \\ \midrule
        \multicolumn{8}{l}{\textbf{This work - zero-shot}} \\
        en-en decoder& 27.3 & 32.2 & 34.0 & 24.7 & 7.4 & 3.3 & 0.1 \\ 
        en-en+noise decoder& 29.2 & 33.3 & 35.3 & 27.3 & 10.1 & 5.2 & 0.3 \\ 
        en-en+de-en decoder& 33.0 & 35.7 & 37.1 & 30.2 & 11.2 & 6.1 & 1.0 \\ \midrule
        \multicolumn{8}{l}{\textbf{Previous work - zero-shot}} \\
        mSLAM \citep{mslam} cross-modal zero-shot & 0.0 & 0.0 & 0.0 & 0.0 & 0.0 & 0.0 & 0.0 \\ \midrule
        \multicolumn{8}{l}{\textbf{Previous works - supervised}} \\
        XLSR (2B) \citep{babu2021xls} & 33.6 & 37.6 & 39.2 & 33.8 & 16.7 & 3.5 & 1.6 \\ 
        mSLAM (2B) \citep{mslam} & 35.9 & 39.0 & 41.0 & 35.4 & 24.2 & 3.3 & 0.8 \\ 
        
         \bottomrule
    \end{tabular}
    \caption{BLEU on CoVoST 2 test set for zero-shot speech-to-text translation (xx $\rightarrow$ en).}
    \label{tab:s2t}
\end{table*}

\paragraph{Text-to-text translation}
As presented in \autoref{sec:archi}, we test different strategies to train an English decoder. When training a decoder with raw text data, we use 15 billion English sentences extracted from CCnet \citep{wenzek2019ccnet}. When training with additional bi-text data, we use bi-texts from CCMatrix \citep{schwenk:2021:acl_ccmatrix}, and the English part of the bi-texts for the auxiliary auto-encoding loss in order to have a good balance between bi-texts and raw data. We present the results for text-to-text translation for xx-en directions in \autoref{tab:t2t_en} for the different decoder training methods on FLORES devtest. \textit{en-en decoder} corresponds to the decoder trained with an auto-encoding objective, \textit{en-en+noise decoder} corresponds to the decoder trained with an auto-encoding objective and additional noise in the sentence embedding space, and \textit{en-en+de-en \mbox{decoder}} corresponds to the decoder trained with a combination of de-en bitexts and english raw data. We compare our zero-shot text-to-text translation results with two supervised baselines: M2M-100 \citep{m2m100}, a massively multilingual trained on many-to-many training data from different sources, with 24 encoder layers and 24 decoder layers; and Deepnet \citep{wang2022deepnet} a recent work trained on 1932 language directions from different sources with 100 encoder layers and 100 decoder layers. We put these results as a supervised reference but we recall that in our framework, we perform zero-shot text-to-text translation for most of the language pairs. Please note the cross-lingual transfer we obtain thanks to our training method: the English decoder has never seen Spanish embeddings before but is able to achieve competitive results compared to supervised baselines.

In \autoref{tab:t2t_en}, we see that adding synthetic noise to the sentence embeddings helps translating low resource languages unseen by the decoder.  However, it slightly decreases the performance on high resource languages. Moreover, natural noise from de-en translations leads to even better results for both high and low resource languages, getting closer to the state-of-the-art MT results which have been obtained with end-to-end training. 

Finally, we trained decoders for German, French, Spanish, Turkish and Mongolian in order to be able to translate from any of our languages to any other. A detailed analysis of the translation tasks with these new decoders can be found in the appendix. Similar to what we noticed with our English decoder, we obtain excellent zero-shot cross-lingual transfer: the German decoder has never seen Japanese embeddings before and Japanese has never been aligned to German. However, the ja-de results are competitive compared to state-of-the-art translation models trained in an end-to-end way with much more data.

\paragraph{Speech-to-text translation}
Then, we tried to plug the decoders trained on text data to our speech encoders in order to perform zero-shot speech-to-text translation. We trained independent speech student encoders for German, French, Turkish, Japanese and Mongolian spoken languages on the \mbox{CoVoST 2} training set. For Catalan and Spanish, we trained a single speech student encoder for both languages as they have high language similarity. We report direct speech translation results in \autoref{tab:s2t} for speech encoders trained with transcriptions as teachers. We have put several baselines for direct speech translation: two supervised baselines based on finetuning XLSR \citep{babu2021xls} or mSLAM \citep{mslam} with speech translation data. We also put the results on zero-shot cross-modal transfer from text to speech with the mSLAM pre-trained multimodal encoder, which is not working in this zero-shot setting.

\begin{table}[t]
    \centering\small
    \begin{tabular}{lllllll}
    \toprule
        %~ &\multicolumn{6}{c}{\textbf{Teacher mode}}\\
        \textbf{Teacher mode:} &\multicolumn{2}{c}{Transcript.} & \multicolumn{2}{c}{Translation} & \multicolumn{2}{c}{Both} \\ 
         & de & ja & de & ja & de & ja \\ \midrule
        en-en & 27.3 & 3.3 & 27.9 & 3.5 & 28.1 & 3.1 \\ 
        en-en+noise & 29.2 & 5.2 & 28.8 & 4.4 & 30.2 & 5.2 \\ 
        en-en+de-en & 33.0 & 6.1 & 30.6 & 4.6 & 33.6 & 5.4 \\ \bottomrule
    \end{tabular}
    \caption{BLEU on CoVoST 2 test set for different teachers and decoders for zero-shot speech-to-text translation.}
    \label{tab:teachers}
\end{table}
\begin{table}[b]
\centering\small
\begin{tabular}{lrr}
\toprule
 & State of the art & Our models \\ \midrule
en$\rightarrow$de & 6.77 & 23.78\\ 
en$\rightarrow$fr & 10.85 & 32.71\\ 
en$\rightarrow$es & 6.75  & 27.43\\ \bottomrule
\end{tabular}
\caption{BLEU on Must-C test set for zero-shot speech translation, compared to the state of the art for zero-shot approaches by \citep{escolano2021enabling}.}
\label{tab:mustc}
\end{table}

\begin{table*}[!ht]
\scalebox{0.89}{
\centering\small
\begin{tabular}{ccc}
\begin{tabular}{lrr}
\toprule
 & es-en & fr-en \\ \midrule
 
 \MC{3}{l}{\textbf{Zero-shot} text-to-speech} \\
 ~~trained on raw speech from CoVoST & 10.0 & 9.5 \\
 ~~trained on raw speech from MLS + Common Voice & 22.8 & 20.9 \\
 ~~trained on \textit{en} ASR data from MLS + Common Voice & 24.4 & 23.5 \\
 \MC{3}{l}{\textbf{Zero-shot} speech-to-speech} \\
  ~~trained on raw speech from CoVoST & 9.9 & 9.1 \\
  ~~trained on raw speech from MLS + Common Voice & 21.3 & 19.8 \\
  ~~trained on \textit{en} ASR data from MLS + Common Voice & 22.4 & 21.1 \\
 \bottomrule
&&\\
\multicolumn{3}{c}{(a) This work: {\em zero-shot} results}\\
&&\\
&&
\end{tabular}%
&~&
\vspace*{-3mm}\begin{tabular}{lrr}
\toprule
  & es-en & fr-en \\ \midrule
 \MC{3}{l}{\textbf{Supervised} speech-to-speech translation} \\ 
~~trained on VP  & 9.2 & 9.6\\ 
~~trained on VP + mined data  & 15.1 & 15.9\\  \midrule
 \MC{3}{l}{\textbf{Supervised} speech-to-speech via text pivot} \\
 ~~trained on VP+EuroparlST+CoVoST & 26.9 & 27.3\\\bottomrule
&&\\

\multicolumn{3}{c}{(b) Results from previous {\em supervised} models trained}\\
\multicolumn{3}{c}{by \citet{s2s_real_data} on real (non synthetic) data.}\\
\multicolumn{3}{c}{The speech-to-speech via text pivot baseline relies}\\
\multicolumn{3}{c}{on speech-to-text by \citet{wang-etal-2021-voxpopuli}.}
\end{tabular}
\end{tabular}
}
\caption{BLEU on CoVoST 2 test set for text-to-speech and speech-to-speech translation}
\label{tab:s2s_results}
\end{table*}

In our framework, the de-en speech translation direction benefits from cross-modal transfer while all other directions benefit from both cross-modal and cross-lingual transfer as the decoder has been trained on text and has only seen English and German embeddings.
In this zero-shot cross-modal setting, we notice that the results are really competitive compared to supervised baselines trained end-to-end. Moreover, the supervised baselines use speech translation data, whereas our approach does not need speech translation data but only transcriptions. Except for Turkish, which has a really different morphological structure compared to English, speech translation results are close to their supervised counterpart trained with XLSR. An interesting direction is ja-en, as we have a large amount of ja-en MT data but a really small amount of speech transcription data. For this task, we nearly doubled the BLEU score compared to supervised baselines without the need of ST data.

We tested the different possible teachers for speech encoder training, namely transcription teacher (already presented), translation teacher, and both transcription and translation teachers. When using translation teacher, we use English text as the written translations from CoVoST~2. We focus on two language directions, de-en (high resource) and ja-en (low resource). Results are shown in \autoref{tab:teachers}.
We notice that a translation teacher is better if using the en-en decoder, which was expected as the decoder was trained on English embeddings. However, when using a decoder trained on noisy embeddings or with additional bi-texts, results are better for speech encoders trained with transcription teacher rather than translation teacher. It may come from the fact that there exists a one-to-one mapping between transcriptions and audios, but not for audio and written translation (there can be several possible translations). For our high resource direction de-en, the best results are achieved when using both transcriptions and translations as teacher, reaching same performance level as with the end-to-end speech translation training of XLSR.

Finally, we trained an English speech student with transcriptions on the Must-C training set and compare our approach with the zero-shot approach by \citet{escolano2021enabling}. We report the results in \autoref{tab:mustc}. 
We notice significant improvements in the BLEU score compared to the previous SOTA for zero-shot speech translation on the Must-C dataset.

\paragraph{Translation of text/speech into speech}
As presented in the \autoref{sec:archi}, we trained English speech decoders with raw English speech only or English speech transcriptions. We present three training settings: one decoder trained on raw English speech data from CoVoST ($\sim$400h), another trained on raw English speech data from both Common Voice ($\sim$2,000h) and Multilingual Librispeech (MLS) ($\sim$40,000h), and finally another trained on English speech transcription data from both Common Voice and Multilingual Librispeech. At test time, we can now plug these English speech decoders to any text or speech encoder. We focused on es-en and fr-en language directions that have previously been covered for direct speech-to-speech translation (see \autoref{tab:s2s_results}).
%We report the results on CoVoST2 test set in Table~\ref{tab:s2s_results} for es-en and fr-en directions.
We also present text-to-speech translation results, plugging text encoders to our speech decoders. 

Following \citet{s2s_paper,s2s_real_data} the evaluation is done by transcribing the output speech with \href{https://huggingface.co/facebook/wav2vec2-large-960h-lv60-self}{an open-sourced ASR system} for English and evaluating the BLEU score of the transcribed speech with target text from CoVoST. We compare these results to a supervised baseline \citep{s2s_real_data} trained on real speech-to-speech translation data from Voxpopuli \citep{wang-etal-2021-voxpopuli} and mined data from \citep{Duquenne:2021:nips_mine}. We also provide a strong supervised baseline composed of a Speech-to-text translation model from \citep{wang-etal-2021-voxpopuli}
that is trained on a significant amount of speech translation data from Voxpopuli, EuroparlST and CoVoST, followed by a text-to-unit model.

In \autoref{tab:s2s_results}, we notice that our speech decoders achieve strong results for this zero-shot setting, even with a limited amount of raw speech data. Incorporating much more raw speech data in the training, significantly improves the results. Using textual representation as input helps in speech decoder training, leading to best results. To the best of our knowledge, these are the first results for zero-shot direct speech-to-speech translation.

This last experiment again highlights the compatibility between representations for different languages and \iffalse different\fi{} modalities. Our approach enables to efficiently leverage raw speech data for T2S and S2S tasks. 

%% file: 5-conclusion.tex
\section{Conclusion}
In this work, we studied how to build a common fixed-size representation for text and speech in different languages, to perform zero-shot cross-modal translation. By imposing a fixed-size representation and aligning explicitly languages and modalities, we have overcome the sentence length mismatch between audio and text, and obtained multilingual and multimodal representations compatible with decoders trained on other languages and/or modalities in a zero-shot setting. We were able to build text and speech encoders for multiple languages compatible with text decoders for multiple languages as well as an English speech decoder. Our zero-shot cross-modal translation results for direct speech-to-text, text-to-speech and speech-to-speech translation define a new zero-shot state-of-the-art baseline. To the best of our knowledge, this is the first work tackling zero-shot direct text-to-speech and speech-to-speech translation. 

Finally, we highlighted the modularity of our architecture; all type of data can be used to train decoders (unlabeled text or speech data ; T2T, S2T, S2S translation data; speech transcription data). Using more types of training data may further enhance the robustness of the decoder to other languages or other modalities.

\section*{Limitations}
We highlighted the modularity of our architecture, learning separately encoders and decoders. While it can be seen as a strength, as one does not need to retrain the whole system to add a new language to the framework, it can also be seen as a limitation as the number of modules increases linearly with the number of languages. Moreover, training multiple separate modules requires more time and computation than one multilingual model. Multilingual training of encoders or decoders is left for future work.

In machine translation, sequence-to-sequence models with fixed-size sentence representation were replaced by sequence-to-sequence models with attention that showed important performance boost for long sentences. Our work shows that competitive performance can still be achieved with fixed-size sentence representations and enables efficient compatibility between languages and modalities. However, very long sequences, beyond usual sentence length, are expected to perform less well.

We showed that it is possible to learn an English speech decoder with raw speech data, it would be interesting to extend this to other languages as target speech, and see how our method performs for a low resource spoken language.

%% file: 6-appendix.tex
\section{Appendix}

\subsection{Distances in LASER text space}
We report the L2 squared distances of sentence embeddings in different languages to their English translations sentence embeddings in LASER space.
\begin{figure}[h!]
    \centering
    \includegraphics[width=0.8\linewidth]{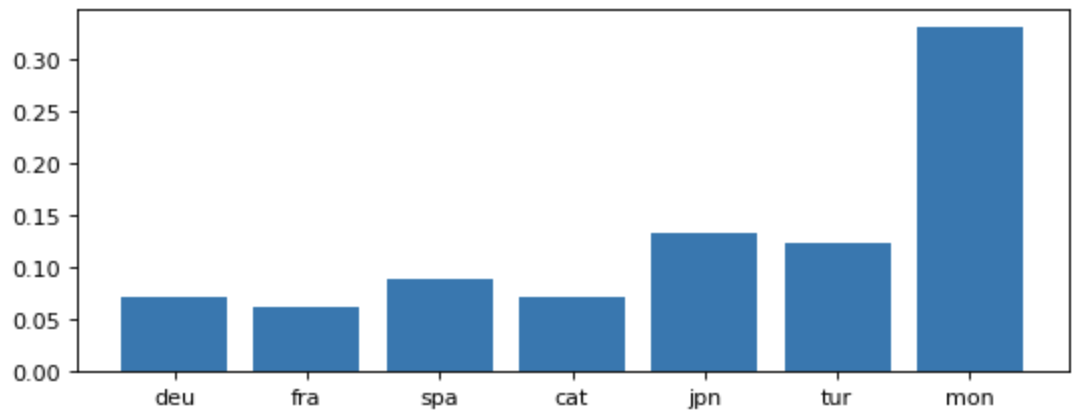}
    \caption{L2 squared distances to English embeddings in LASER space for translations from FLORES devtest}
    \label{fig:distances}
\end{figure}

\subsection{Other text decoders}
With the conclusion that bi-text data can help the decoder be robust to other unseen languages, we trained decoders for German, French, Spanish, Turkish and Mongolian. We use en-xx bitexts, in addition to raw xx data to train the decoders. For all decoder trainings, we use bi-texts from CCMatrix \citep{schwenk:2021:acl_ccmatrix}, for the auto-encoding loss we use one side of the bi-texts corresponding to the language that we are trying to decode, except for Mongolian where we take all the raw Mongolian text data from CCnet.  \citep{wenzek2019ccnet}. We present the results in Table~\ref{tab:t2t_xx}.  

\begin{table*}[!ht]
    \centering\small
    \begin{tabular}{l*{8}{r}}
    
    \toprule
         & \MC{1}{c}{en} & \MC{1}{c}{de} & \MC{1}{c}{fr} & \MC{1}{c}{es} & \MC{1}{c}{ca} & \MC{1}{c}{ja} & \MC{1}{c}{tr} & \MC{1}{c}{mn} \\ \midrule
        \multicolumn{9}{l}{\textbf{Translation into German}} \\
        \multicolumn{9}{l}{~~~\textbf{\textit{This work - zero-shot expect for en-de}}} \\
        ~~~de-de+en-de decoder & 39.1 & --- & 32.6 & 24.6 & 29.2 & 20.9 & 27.9 & 12.8 \\ 
        \multicolumn{9}{l}{~~~\textbf{\textit{Previous works - supervised}}} \\
        ~~~M2M-100 (12B - 48 layers) \citep{m2m100} & 42.1 & --- & 34.5 & 27.1 & 30.9 & 21.4 & 28.4 & 15.9 \\ 
        ~~~Deepnet (3.2B - 200 layers) \citep{wang2022deepnet} & 46.0 & --- & 36.2 & 29.2 & 32.5 & 24.7 & 31.9 & 21.7 \\ \midrule
        \multicolumn{9}{l}{\textbf{Translation into Spanish}} \\
        \multicolumn{9}{l}{~~~\textbf{\textit{This work - zero-shot expect for en-es}}} \\
        ~~~es-es+en-es decoder & 29.1 & 25.9 & 26.8 & --- & 26.3 & 18.6 & 22.8 & 12.2  \\
        \multicolumn{9}{l}{~~~\textbf{\textit{Previous works - supervised}}} \\
        ~~~M2M-100 (12B - 48 layers) \citep{m2m100} & 30.3 & 27.2 & 28.2 & --- & 26.6 & 19.4 & 24.0 & 14.9 \\
        ~~~Deepnet (3.2B - 200 layers) \citep{wang2022deepnet} & 32.2 & 28.3 & 28.8 & --- & 26.9 & 21.5 & 25.9 & 18.8
        \\\midrule
        \multicolumn{9}{l}{\textbf{Translation into French}} \\
        \multicolumn{9}{l}{~~~\textbf{\textit{This work - zero-shot expect for en-fr}}} \\
        ~~~fr-fr+en-fr decoder  & 49.1 & 38.3 & --- & 31.2 & 37.6 & 25.3 & 33.4 & 16.6 \\
        \multicolumn{9}{l}{~~~\textbf{\textit{Previous works - supervised}}} \\
        ~~~M2M-100 (12B - 48 layers) \citep{m2m100} & 51.4 & 42 & --- & 32.8 & 39.7 & 26.6 & 35.1 & 20.8 \\
        ~~~Deepnet (3.2B - 200 layers) \citep{wang2022deepnet} & 54.7 & 43.4 & --- & 35.2 & 41.6 & 29.9 & 38.2 & 26.6 \\
        \midrule
        \multicolumn{9}{l}{\textbf{Translation into Turkish}} \\
        \multicolumn{9}{l}{~~~\textbf{\textit{This work - zero-shot expect for en-tr}}} \\ 
        ~~~tr-tr+en-tr decoder & 31.2 & 27.1 & 26.4 & 21.5 & 24.2 & 19.1 & --- & 13.7 \\
        \multicolumn{9}{l}{~~~\textbf{\textit{Previous works - supervised}}} \\
        ~~~M2M-100 (12B - 48 layers) \citep{m2m100} & 32.8 & 26.9 & 26.6 & 22.3 & 24.3 & 18.6 & --- & 16.1 \\
        ~~~Deepnet (3.2B - 200 layers) \citep{wang2022deepnet} & 39.5 & 32.0 & 31.6 & 26.2 & 28.2 & 23.2 & --- & 21.0 \\\midrule
        \multicolumn{9}{l}{\textbf{Translation into Mongolian}} \\
        \multicolumn{9}{l}{~~~\textbf{\textit{This work - zero-shot expect for en-mn}}} \\
        ~~~mn-mn+en-mn decoder & 15.7 & 15.8 & 15.2 & 13.6 & 15.2 & 13.5 & 15.4 & --- \\
        \multicolumn{9}{l}{~~~\textbf{\textit{Previous works - supervised}}} \\
        ~~~M2M-100 (12B - 48 layers) \citep{m2m100} & 12.0 & 10.7 & 10.9 & 9.2 & 10.8 & 9.3 & 11.0 & --- \\
        ~~~Deepnet (3.2B - 200 layers) \citep{wang2022deepnet} & 18.3 & 16.8 & 16.2 & 15.0 & 15.8 & 13.7 & 15.9 & --- \\ \bottomrule
    \end{tabular}
    \caption{BLEU on FLORES devtest for text-to-text translation for de, es, fr, tr and mn decoders}
     \label{tab:t2t_xx}
\end{table*}

\subsection{Training details}
We use Fairseq to train our models. Text student encoders are trained on 32 Tesla V100 GPUs with a learning rate set to $10^{-4}$, maximum number of tokens by GPU is 1400, and update frequency is set to 4. Speech student encoders are trained on 48 Tesla V100 GPUs for a few days, with same learning rate as text students, maximum number of sentences is set to 32 by GPU. Text decoders are trained with the same configuration as mBART. Speech decoders are trained on 48 Tesla V100 GPUs with a learning rate set to $3 \cdot 10^{-4}$, maximum number of sentences is set to 32 by GPU and update frequency is set to~4.

%% file: emnlp2022.bbl
\begin{thebibliography}{47}
\expandafter\ifx\csname natexlab\endcsname\relax\def\natexlab#1{#1}\fi

\bibitem[{Alinejad and Sarkar(2020)}]{alinejad2020effectively}
Ashkan Alinejad and Anoop Sarkar. 2020.
\newblock Effectively pretraining a speech translation decoder with machine
  translation data.
\newblock In \emph{Proceedings of the 2020 Conference on Empirical Methods in
  Natural Language Processing (EMNLP)}, pages 8014--8020.

\bibitem[{Arivazhagan et~al.(2019)Arivazhagan, Bapna, Firat, Aharoni, Johnson,
  and Macherey}]{arivazhagan2019missing}
Naveen Arivazhagan, Ankur Bapna, Orhan Firat, Roee Aharoni, Melvin Johnson, and
  Wolfgang Macherey. 2019.
\newblock The missing ingredient in zero-shot neural machine translation.
\newblock \emph{arXiv preprint arXiv:1903.07091}.

\bibitem[{Artetxe and Schwenk(2019)}]{Artetxe:2019:tacl_massive_ml}
Mikel Artetxe and Holger Schwenk. 2019.
\newblock Massively multilingual sentence embeddings for zero-shot
  cross-lingual transfer and beyond.
\newblock \emph{TACL}, pages 597--610.

\bibitem[{Babu et~al.(2021)Babu, Wang, Tjandra, Lakhotia, Xu, Goyal, Singh, von
  Platen, Saraf, Pino et~al.}]{babu2021xls}
Arun Babu, Changhan Wang, Andros Tjandra, Kushal Lakhotia, Qiantong Xu, Naman
  Goyal, Kritika Singh, Patrick von Platen, Yatharth Saraf, Juan Pino, et~al.
  2021.
\newblock Xls-r: Self-supervised cross-lingual speech representation learning
  at scale.
\newblock \emph{arXiv preprint arXiv:2111.09296}.

\bibitem[{Baevski et~al.(2022)Baevski, Hsu, Xu, Babu, Gu, and
  Auli}]{baevski:2022:arxiv_data2vec}
Alexei Baevski, Wei{-}Ning Hsu, Qiantong Xu, Arun Babu, Jiatao Gu, and Michael
  Auli. 2022.
\newblock data2vec: {A} general framework for self-supervised learning in
  speech, vision and language.
\newblock In \emph{\url{https://arxiv.org/abs/2202.03555}}.

\bibitem[{Baevski et~al.(2020)Baevski, Zhou, Mohamed, and Auli}]{wav2vec2}
Alexei Baevski, Yuhao Zhou, Abdelrahman Mohamed, and Michael Auli. 2020.
\newblock wav2vec 2.0: A framework for self-supervised learning of speech
  representations.
\newblock \emph{Advances in Neural Information Processing Systems},
  33:12449--12460.

\bibitem[{Ba{\~n}{\'o}n et~al.(2020)Ba{\~n}{\'o}n, Chen, Haddow, Heafield,
  Hoang, Espl{\`a}-Gomis, Forcada, Kamran, Kirefu, Koehn, Ortiz~Rojas,
  Pla~Sempere, Ram{\'\i}rez-S{\'a}nchez, Sarr{\'\i}as, Strelec, Thompson,
  Waites, Wiggins, and Zaragoza}]{banon-etal-2020-paracrawl}
Marta Ba{\~n}{\'o}n, Pinzhen Chen, Barry Haddow, Kenneth Heafield, Hieu Hoang,
  Miquel Espl{\`a}-Gomis, Mikel~L. Forcada, Amir Kamran, Faheem Kirefu, Philipp
  Koehn, Sergio Ortiz~Rojas, Leopoldo Pla~Sempere, Gema
  Ram{\'\i}rez-S{\'a}nchez, Elsa Sarr{\'\i}as, Marek Strelec, Brian Thompson,
  William Waites, Dion Wiggins, and Jaume Zaragoza. 2020.
\newblock \href {https://doi.org/10.18653/v1/2020.acl-main.417} {{P}ara{C}rawl:
  Web-scale acquisition of parallel corpora}.
\newblock In \emph{Proceedings of the 58th Annual Meeting of the Association
  for Computational Linguistics}, pages 4555--4567, Online. Association for
  Computational Linguistics.

\bibitem[{Bansal et~al.(2017)Bansal, Kamper, Lopez, and
  Goldwater}]{bansal2017towards}
Sameer Bansal, Herman Kamper, Adam Lopez, and Sharon Goldwater. 2017.
\newblock Towards speech-to-text translation without speech recognition.
\newblock \emph{arXiv preprint arXiv:1702.03856}.

\bibitem[{Bapna et~al.(2022)Bapna, Cherry, Zhang, Jia, Johnson, Cheng, Khanuja,
  Riesa, and Conneau}]{mslam}
Ankur Bapna, Colin Cherry, Yu~Zhang, Ye~Jia, Melvin Johnson, Yong Cheng, Simran
  Khanuja, Jason Riesa, and Alexis Conneau. 2022.
\newblock mslam: Massively multilingual joint pre-training for speech and text.
\newblock \emph{arXiv preprint arXiv:2202.01374}.

\bibitem[{B{\'e}rard et~al.(2016)B{\'e}rard, Pietquin, Servan, and
  Besacier}]{berard2016listen}
Alexandre B{\'e}rard, Olivier Pietquin, Christophe Servan, and Laurent
  Besacier. 2016.
\newblock Listen and translate: A proof of concept for end-to-end
  speech-to-text translation.
\newblock \emph{arXiv preprint arXiv:1612.01744}.

\bibitem[{Conneau et~al.(2020)Conneau, Baevski, Collobert, Mohamed, and
  Auli}]{xlsr}
Alexis Conneau, Alexei Baevski, Ronan Collobert, Abdelrahman Mohamed, and
  Michael Auli. 2020.
\newblock Unsupervised cross-lingual representation learning for speech
  recognition.
\newblock \emph{arXiv preprint arXiv:2006.13979}.

\bibitem[{Conneau and Lample(2019)}]{xlm}
Alexis Conneau and Guillaume Lample. 2019.
\newblock Cross-lingual language model pretraining.
\newblock \emph{Advances in neural information processing systems}, 32.

\bibitem[{Di~Gangi et~al.(2019)Di~Gangi, Cattoni, Bentivogli, Negri, and
  Turchi}]{mustc:2019:naacl}
Mattia~A. Di~Gangi, Roldano Cattoni, Luisa Bentivogli, Matteo Negri, and Marco
  Turchi. 2019.
\newblock \href {https://aclanthology.org/N19-1202} {{M}u{ST}-{C}: a
  {M}ultilingual {S}peech {T}ranslation {C}orpus}.
\newblock In \emph{NAACL}, pages 2012--2017.

\bibitem[{Dinh(2021)}]{dinh2021zero}
Tu~Anh Dinh. 2021.
\newblock Zero-shot speech translation.
\newblock \emph{arXiv preprint arXiv:2107.06010}.

\bibitem[{Dinh et~al.(2022)Dinh, Liu, and Niehues}]{dinh2022tackling}
Tu~Anh Dinh, Danni Liu, and Jan Niehues. 2022.
\newblock Tackling data scarcity in speech translation using zero-shot
  multilingual machine translation techniques.
\newblock \emph{arXiv preprint arXiv:2201.11172}.

\bibitem[{Dong et~al.(2021)Dong, Ye, Wang, Zhou, Xu, Xu, and
  Li}]{dong2021listen}
Qianqian Dong, Rong Ye, Mingxuan Wang, Hao Zhou, Shuang Xu, Bo~Xu, and Lei Li.
  2021.
\newblock Listen, understand and translate: Triple supervision decouples
  end-to-end speech-to-text translation.
\newblock In \emph{Proceedings of the AAAI Conference on Artificial
  Intelligence}, volume~35, pages 12749--12759.

\bibitem[{Duquenne et~al.(2021)Duquenne, Gong, and
  Schwenk}]{Duquenne:2021:nips_mine}
Paul-Ambroise Duquenne, Hongyu Gong, and Holger Schwenk. 2021.
\newblock Multimodal and multilingual embeddings for large-scale speech mining.
\newblock \emph{Advances in Neural Information Processing Systems}, 34.

\bibitem[{Escolano et~al.(2021{\natexlab{a}})Escolano, Costa-Juss{\`a}, and
  Fonollosa}]{escolano2021bilingual}
Carlos Escolano, Marta~R Costa-Juss{\`a}, and Jos{\'e}~AR Fonollosa.
  2021{\natexlab{a}}.
\newblock From bilingual to multilingual neural-based machine translation by
  incremental training.
\newblock \emph{Journal of the Association for Information Science and
  Technology}, 72(2):190--203.

\bibitem[{Escolano et~al.(2020{\natexlab{a}})Escolano, Costa-juss{\`a},
  Fonollosa, and Artetxe}]{escolano2020multilingual}
Carlos Escolano, Marta~R Costa-juss{\`a}, Jos{\'e}~AR Fonollosa, and Mikel
  Artetxe. 2020{\natexlab{a}}.
\newblock Multilingual machine translation: Closing the gap between shared and
  language-specific encoder-decoders.
\newblock \emph{arXiv preprint arXiv:2004.06575}.

\bibitem[{Escolano et~al.(2020{\natexlab{b}})Escolano, Costa-juss{\`a},
  Fonollosa, and Artetxe}]{escolano2020training}
Carlos Escolano, Marta~R Costa-juss{\`a}, Jos{\'e}~AR Fonollosa, and Mikel
  Artetxe. 2020{\natexlab{b}}.
\newblock Training multilingual machine translation by alternately freezing
  language-specific encoders-decoders.
\newblock \emph{arXiv preprint arXiv:2006.01594}.

\bibitem[{Escolano et~al.(2021{\natexlab{b}})Escolano, Costa-juss{\`a},
  Fonollosa, and Segura}]{escolano2021enabling}
Carlos Escolano, Marta~R Costa-juss{\`a}, Jos{\'e}~AR Fonollosa, and Carlos
  Segura. 2021{\natexlab{b}}.
\newblock Enabling zero-shot multilingual spoken language translation with
  language-specific encoders and decoders.
\newblock In \emph{2021 IEEE Automatic Speech Recognition and Understanding
  Workshop (ASRU)}, pages 694--701. IEEE.

\bibitem[{Fan et~al.(2021)Fan, Bhosale, Schwenk, Ma, El-Kishky, Goyal, Baines,
  Celebi, Wenzek, Chaudhary et~al.}]{m2m100}
Angela Fan, Shruti Bhosale, Holger Schwenk, Zhiyi Ma, Ahmed El-Kishky,
  Siddharth Goyal, Mandeep Baines, Onur Celebi, Guillaume Wenzek, Vishrav
  Chaudhary, et~al. 2021.
\newblock Beyond english-centric multilingual machine translation.
\newblock \emph{Journal of Machine Learning Research}, 22(107):1--48.

\bibitem[{Feng et~al.(2020)Feng, Yang, Cer, Arivazhagan, and Wang}]{labse}
Fangxiaoyu Feng, Yinfei Yang, Daniel Cer, Naveen Arivazhagan, and Wei Wang.
  2020.
\newblock Language-agnostic bert sentence embedding.
\newblock \emph{arXiv preprint arXiv:2007.01852}.

\bibitem[{Goyal et~al.(2022)Goyal, Gao, Chaudhary, Chen, Wenzek, Ju, Krishnan,
  Ranzato, Guzman, and Fan}]{flores}
Naman Goyal, Cynthia Gao, Vishrav Chaudhary, Peng-Jen Chen, Guillaume Wenzek,
  Da~Ju, Sanjana Krishnan, Marc’Aurelio Ranzato, Francisco Guzman, and Angela
  Fan. 2022.
\newblock The flores-101 evaluation benchmark for low-resource and multilingual
  machine translation.
\newblock \emph{Transactions of the Association for Computational Linguistics},
  10:522--538.

\bibitem[{Heffernan et~al.(2022)Heffernan, {\c{C}}elebi, and Schwenk}]{laser3}
Kevin Heffernan, Onur {\c{C}}elebi, and Holger Schwenk. 2022.
\newblock Bitext mining using distilled sentence representations for
  low-resource languages.
\newblock \emph{arXiv preprint arXiv:2205.12654}.

\bibitem[{Jia et~al.(2021)Jia, Ramanovich, Remez, and
  Pomerantz}]{jia2021translatotron}
Ye~Jia, Michelle~Tadmor Ramanovich, Tal Remez, and Roi Pomerantz. 2021.
\newblock Translatotron 2: Robust direct speech-to-speech translation.
\newblock \emph{arXiv preprint arXiv:2107.08661}.

\bibitem[{Jia et~al.(2019)Jia, Weiss, Biadsy, Macherey, Johnson, Chen, and
  Wu}]{jia2019direct}
Ye~Jia, Ron~J Weiss, Fadi Biadsy, Wolfgang Macherey, Melvin Johnson, Zhifeng
  Chen, and Yonghui Wu. 2019.
\newblock Direct speech-to-speech translation with a sequence-to-sequence
  model.
\newblock \emph{arXiv preprint arXiv:1904.06037}.

\bibitem[{Lee et~al.(2022{\natexlab{a}})Lee, Chen, Wang, Gu, Popuri, Ma,
  Polyak, Adi, He, Tang, Pino, and Hsu}]{s2s_paper}
Ann Lee, Peng-Jen Chen, Changhan Wang, Jiatao Gu, Sravya Popuri, Xutai Ma, Adam
  Polyak, Yossi Adi, Qing He, Yun Tang, Juan Pino, and Wei-Ning Hsu.
  2022{\natexlab{a}}.
\newblock \href {https://aclanthology.org/2022.acl-long.235} {Direct
  speech-to-speech translation with discrete units}.
\newblock In \emph{ACL}, pages 3327--3339.

\bibitem[{Lee et~al.(2022{\natexlab{b}})Lee, Gong, Duquenne, Schwenk, Chen,
  Wang, Popuri, Adi, Pino, Gu, and Hsu}]{s2s_real_data}
Ann Lee, Hongyu Gong, Paul-Ambroise Duquenne, Holger Schwenk, Peng-Jen Chen,
  Changhan Wang, Sravya Popuri, Yossi Adi, Juan Pino, Jiatao Gu, and Wei-Ning
  Hsu. 2022{\natexlab{b}}.
\newblock \href {https://aclanthology.org/2022.naacl-main.63} {Textless
  speech-to-speech translation on real data}.
\newblock In \emph{NAACL}, pages 860--872.

\bibitem[{Li et~al.(2020)Li, Wang, Tang, Tran, Tang, Pino, Baevski, Conneau,
  and Auli}]{li2020multilingual}
Xian Li, Changhan Wang, Yun Tang, Chau Tran, Yuqing Tang, Juan Pino, Alexei
  Baevski, Alexis Conneau, and Michael Auli. 2020.
\newblock Multilingual speech translation with efficient finetuning of
  pretrained models.
\newblock \emph{arXiv preprint arXiv:2010.12829}.

\bibitem[{Liao et~al.(2021)Liao, Shi, Gong, Shou, Qu, and
  Zeng}]{liao2021improving}
Junwei Liao, Yu~Shi, Ming Gong, Linjun Shou, Hong Qu, and Michael Zeng. 2021.
\newblock Improving zero-shot neural machine translation on language-specific
  encoders-decoders.
\newblock In \emph{2021 International Joint Conference on Neural Networks
  (IJCNN)}, pages 1--8. IEEE.

\bibitem[{Liu et~al.(2020)Liu, Zhu, Zhang, and Zong}]{liu2020bridging}
Yuchen Liu, Junnan Zhu, Jiajun Zhang, and Chengqing Zong. 2020.
\newblock Bridging the modality gap for speech-to-text translation.
\newblock \emph{arXiv preprint arXiv:2010.14920}.

\bibitem[{Lu et~al.(2018)Lu, Keung, Ladhak, Bhardwaj, Zhang, and
  Sun}]{lu2018neural}
Yichao Lu, Phillip Keung, Faisal Ladhak, Vikas Bhardwaj, Shaonan Zhang, and
  Jason Sun. 2018.
\newblock A neural interlingua for multilingual machine translation.
\newblock \emph{arXiv preprint arXiv:1804.08198}.

\bibitem[{Pham et~al.(2019)Pham, Niehues, Ha, and Waibel}]{pham2019improving}
Ngoc-Quan Pham, Jan Niehues, Thanh-Le Ha, and Alex Waibel. 2019.
\newblock Improving zero-shot translation with language-independent
  constraints.
\newblock \emph{arXiv preprint arXiv:1906.08584}.

\bibitem[{Reimers and Gurevych(2019)}]{sentence_bert}
Nils Reimers and Iryna Gurevych. 2019.
\newblock Sentence-bert: Sentence embeddings using siamese bert-networks.
\newblock \emph{arXiv preprint arXiv:1908.10084}.

\bibitem[{Reimers and Gurevych(2020)}]{Reimers}
Nils Reimers and Iryna Gurevych. 2020.
\newblock Making monolingual sentence embeddings multilingual using knowledge
  distillation.
\newblock In \emph{EMNLP}, pages 4512--4525.

\bibitem[{Salesky et~al.(2021)Salesky, Wiesner, Bremerman, Cattoni, Negri,
  Turchi, Oard, and Post}]{salesky2021multilingual}
Elizabeth Salesky, Matthew Wiesner, Jacob Bremerman, Roldano Cattoni, Matteo
  Negri, Marco Turchi, Douglas~W Oard, and Matt Post. 2021.
\newblock The multilingual tedx corpus for speech recognition and translation.
\newblock \emph{arXiv preprint arXiv:2102.01757}.

\bibitem[{Schwenk et~al.(2021)Schwenk, Wenzek, Edunov, Grave, Joulin, and
  Fan}]{schwenk:2021:acl_ccmatrix}
Holger Schwenk, Guillaume Wenzek, Sergey Edunov, Edouard Grave, Armand Joulin,
  and Angela Fan. 2021.
\newblock {CCMatrix}: Mining billions of high-quality parallel sentences on the
  web.
\newblock In \emph{ACL}, page 6490–6500.

\bibitem[{V{\'a}zquez et~al.(2018)V{\'a}zquez, Raganato, Tiedemann, and
  Creutz}]{vazquez2018multilingual}
Ra{\'u}l V{\'a}zquez, Alessandro Raganato, J{\"o}rg Tiedemann, and Mathias
  Creutz. 2018.
\newblock Multilingual nmt with a language-independent attention bridge.
\newblock \emph{arXiv preprint arXiv:1811.00498}.

\bibitem[{Wang et~al.(2020{\natexlab{a}})Wang, Pino, Wu, and
  Gu}]{covost1:2020:lrec}
Changhan Wang, Juan Pino, Anne Wu, and Jiatao Gu. 2020{\natexlab{a}}.
\newblock \href {https://www.aclweb.org/anthology/2020.lrec-1.517}
  {{C}o{V}o{ST}: A diverse multilingual speech-to-text translation corpus}.
\newblock In \emph{LREC}, pages 4197--4203.

\bibitem[{Wang et~al.(2021)Wang, Riviere, Lee, Wu, Talnikar, Haziza,
  Williamson, Pino, and Dupoux}]{wang-etal-2021-voxpopuli}
Changhan Wang, Morgane Riviere, Ann Lee, Anne Wu, Chaitanya Talnikar, Daniel
  Haziza, Mary Williamson, Juan Pino, and Emmanuel Dupoux. 2021.
\newblock \href {https://aclanthology.org/2021.acl-long.80} {{V}ox{P}opuli: A
  large-scale multilingual speech corpus for representation learning,
  semi-supervised learning and interpretation}.
\newblock In \emph{Proceedings of the 59th Annual Meeting of the Association
  for Computational Linguistics and the 11th International Joint Conference on
  Natural Language Processing (Volume 1: Long Papers)}, pages 993--1003,
  Online. Association for Computational Linguistics.

\bibitem[{Wang et~al.(2020{\natexlab{b}})Wang, Wu, and
  Pino}]{covost2:2020:arxiv}
Changhan Wang, Anne Wu, and Juan Pino. 2020{\natexlab{b}}.
\newblock \href {http://arxiv.org/abs/2007.10310} {{CoVoST 2}: A massively
  multilingual speech-to-text translation corpus}.

\bibitem[{Wang et~al.(2022)Wang, Ma, Dong, Huang, Zhang, and
  Wei}]{wang2022deepnet}
Hongyu Wang, Shuming Ma, Li~Dong, Shaohan Huang, Dongdong Zhang, and Furu Wei.
  2022.
\newblock Deepnet: Scaling transformers to 1,000 layers.
\newblock \emph{arXiv preprint arXiv:2203.00555}.

\bibitem[{Weiss et~al.(2017)Weiss, Chorowski, Jaitly, Wu, and
  Chen}]{weiss2017sequence}
Ron~J Weiss, Jan Chorowski, Navdeep Jaitly, Yonghui Wu, and Zhifeng Chen. 2017.
\newblock Sequence-to-sequence models can directly translate foreign speech.
\newblock \emph{arXiv preprint arXiv:1703.08581}.

\bibitem[{Wenzek et~al.(2019)Wenzek, Lachaux, Conneau, Chaudhary, Guzm{\'a}n,
  Joulin, and Grave}]{wenzek2019ccnet}
Guillaume Wenzek, Marie-Anne Lachaux, Alexis Conneau, Vishrav Chaudhary,
  Francisco Guzm{\'a}n, Armand Joulin, and Edouard Grave. 2019.
\newblock Ccnet: Extracting high quality monolingual datasets from web crawl
  data.
\newblock \emph{arXiv preprint arXiv:1911.00359}.

\bibitem[{Xu et~al.(2021)Xu, Hu, Li, Zhang, Ju, Xiao, Zhu
  et~al.}]{xu2021stacked}
Chen Xu, Bojie Hu, Yanyang Li, Yuhao Zhang, Qi~Ju, Tong Xiao, Jingbo Zhu,
  et~al. 2021.
\newblock Stacked acoustic-and-textual encoding: Integrating the pre-trained
  models into speech translation encoders.
\newblock \emph{arXiv preprint arXiv:2105.05752}.

\bibitem[{Zhang et~al.(2022)Zhang, Li, and Liu}]{zhang2022triangular}
Meng Zhang, Liangyou Li, and Qun Liu. 2022.
\newblock Triangular transfer: Freezing the pivot for triangular machine
  translation.
\newblock \emph{arXiv preprint arXiv:2203.09027}.

\end{thebibliography}
